# Model-agnostic explainable artificial intelligence for object detection in image data

Milad Moradi[*]

AI Research, Tricentis, Vienna, Austria

m.moradi-vastegani@tricentis.com

Ke Yan

AI Research, Tricentis, Sydney, Australia

k.yan@tricentis.com

David Colwell

AI Research, Tricentis, Sydney, Australia

d.colwell@tricentis.com

Matthias Samwald

Institute of Artificial Intelligence, Center for Medical Statistics, Informatics, and Intelligent Systems, Medical University of Vienna, Vienna, Austria

matthias.samwald@meduniwien.ac.at

Rhona Asgari

AI Research, Tricentis, Vienna, Austria

r.asgari@tricentis.com

---

[*] Corresponding author. **Postal address:** Tricentis GmbH, Leonard-Bernstein-Straße 10, 1220 Vienna, Austria.

**Abstract**

In recent years, deep neural networks have been widely used for building high-performance Artificial Intelligence (AI) systems for computer vision applications. Object detection is a fundamental task in computer vision, which has been greatly progressed through developing large and intricate AI models. However, the lack of transparency is a big challenge that may not allow the widespread adoption of these models. Explainable artificial intelligence is a field of research where methods are developed to help users understand the behavior, decision logics, and vulnerabilities of AI systems. Previously, few explanation methods were developed for object detection based on random masking. However, random masks may raise some issues regarding the actual importance of pixels within an image. In this paper, we design and implement a black-box explanation method named **B**lack-box **O**bject **D**etection **E**xplanation by **M**asking (BODEM) through adopting a hierarchical random masking approach for object detection systems. We propose a hierarchical random masking framework in which coarse-grained masks are used in lower levels to find salient regions within an image, and fine-grained mask are used to refine the salient regions in higher levels. Experimentations on various object detection datasets and models showed that BODEM can effectively explain the behavior of object detectors. Moreover, our method outperformed Detector Randomized Input Sampling for Explanation (D-RISE) and Local Interpretable Model-agnostic Explanations (LIME) with respect to different quantitative measures of explanation effectiveness. The experimental results demonstrate that BODEM can be an effective method for explaining and validating object detection systems in black-box testing scenarios.

# 1. Introduction and background

Artificial Intelligence (AI) and Machine Learning (ML) methods have seen extensive use in recent years for developing intelligent systems capable of performing tasks that typically require human intelligence [1-5]. A significant factor in the success of AI systems is the use of Deep Neural Networks (DNNs), which excel in capturing complex data relationships. When trained on sufficiently large datasets, DNNs can function as either high-performance generative or discriminative ML models [6].

A DNN features a hierarchical architecture of layers, with each layer comprising multiple non-linear processing units. During training, the lower layers (closer to the input) learn simpler data relationships and pass them on to subsequent layers. As the input data progresses through the DNN layers, it captures increasingly complex patterns, with each layer building on the simpler patterns identified by previous layers. The upper layers, closer to the output, typically act as functions that approximate the target output specified by the task. Common DNN architectures include multi-layer perceptrons, deep belief networks, autoencoders, Convolutional Neural Networks (CNNs), Recurrent Neural Networks (RNNs), and transformers, each with applications in image, text, time series, and tabular data processing [2, 6, 7].

Among these architectures, CNNs are particularly popular. Inspired by the natural visual perception of living creatures, CNNs are ideal for extracting patterns from multi-dimensional data with a grid-like topology, such as images and videos [8]. Convolutional layers in a CNN apply trainable filters to the input at all possible receptive fields, producing local feature maps. Pooling layers then subsample these feature maps to reduce their size. As the input passes through multiple convolutional and pooling layers, higher-level transformations of the input are constructed. After rasterizing the final feature map into a one-dimensional vector, the output is fed into one or more fully-connected layers to generate the final result.

CNNs have diverse applications in computer vision and image processing, aiding AI systems in deriving high-level understanding from digital images and detecting complex structures [8, 9]. A CNN consists of simple yet non-linear modules that learn multiple levels of abstraction through its layers. The input to the first layer typically consists of arrays of pixel values across different color channels, such as Red, Green, and Blue (RGB). The initial layer usually detects the presence or absence of edges with particular orientations at specific locations in the image. Subsequent layers detect combinations of these edges and encode motifs into feature maps, which then identify parts of familiar objects by combining motifs. Finally, the last layers assemble these parts to detect whole objects [2].

## 1.1. Object detection

Object detection is a fundamental task in image processing aimed at localizing and classifying objects within an image [10]. This process provides critical information for the semantic understanding of images and videos, with applications spanning autonomous driving, image classification, face recognition, and other related fields [11]. The advent of DNNs has significantly enhanced the performance of computer vision applications, particularly in object detection. CNNs typically form the backbone of object detection systems, serving as feature detectors [12]. Prominent object detection models include Regions with CNN (R-CNN) [13], Fast R-CNN [14], Faster R-CNN [15], Mask R-CNN [16], You Only Look Once (YOLO) [17], and Single-Shot Detector (SSD) [18].

The advancements in object detection methods have led to their widespread application in various real-world scenarios, transforming numerous industries and enhancing technological capabilities. In autonomous driving, for example, accurate and real-time object detection is crucial for identifying pedestrians, other vehicles, and obstacles, thereby ensuring safe navigation and collision avoidance [19, 20]. Car manufacturing companies utilize sophisticated object detection models to enable their self-driving cars to interpret and react to complex environments. In the field of surveillance and security, object detection systems are employed to monitor and analyze video feeds, detecting unauthorized access or identifying suspicious behavior in real-time, which is instrumental in preventing security breaches and managing public safety [21-23].

In software engineering, object detection methods are utilized in automated software testing to identify and interact with user interface elements, ensuring that applications function correctly across different platforms and devices [24]. These techniques help streamline the testing process by automatically recognizing and validating user interface components, reducing the need for manual intervention and increasing testing efficiency. Healthcare also benefits significantly from these technologies, where medical imaging systems leverage object detection to identify anomalies such as tumors in radiographs and Magnetic Resonance Images (MRIs), aiding in early diagnosis and treatment planning [25-27]. Moreover, in the retail industry, object detection is used for automated inventory management and checkout systems, enhancing efficiency and customer experience [28, 29]. These applications demonstrate the profound impact of state-of-the-art object detection methods across diverse sectors, driving innovation and improving operational effectiveness.

Despite the impressive success of Deep Learning (DL) methods, their lack of interpretability remains a significant barrier to widespread adoption, especially in mission-critical applications [30]. For users to trust an ML model, they must be able to explain or interpret what the model has learned [31]. A clear understanding of an ML model's decisions, potential biases, and vulnerabilities facilitates trust, enables the detection of biases, informs actions to refine the model, and optimizes its performance [32-34].

## 1.2. Explainable artificial intelligence

Explainable AI (XAI) is a research field focused on developing methods that enable users to understand, explain, and interpret AI and ML systems [35]. Several studies have revealed that decisions made by black-box AI models were biased or unfair upon examination through XAI techniques [30]. In computer vision, XAI has been instrumental in identifying biases and failure points in DL models used for object detection and classification [36, 37]. XAI methods for computer vision are generally categorized into white-box and black-box approaches, based on their access to the DL model's internals.

White-box methods have access to information about the DL network, such as its architecture, loss function, activation functions, connection weights, and training data. Conversely, black-box XAI methods operate without access to this internal information, relying solely on the input sample and the final output generated by the DL model. A significant challenge of white-box methods is the frequent unavailability of the model's internal details. Additionally, interpreting explanations derived from the model's internals often requires expertise in AI, limiting their accessibility. In scenarios where understanding the AI model's behaviour and discovering its vulnerabilities is crucial, such as in software testing [38], black-box explanations are often more practical.

XAI methods for computer vision tasks aim to make the decision-making process of AI systems transparent and interpretable. These methods are critical for applications where understanding the rationale behind a model's predictions is essential, such as in medical imaging, autonomous driving, software quality assurance, and security surveillance. XAI techniques for computer vision include saliency maps [39], which highlight the regions in an image that contribute most to the model's decision, and Class Activation Maps (CAMs) [40], which provide a visual explanation of which parts of an image are responsible for activating certain classes in a convolutional neural network. Additionally, perturbation-based methods systematically alter parts of an image to observe changes in the model's output, thus identifying significant features. These approaches help users trust and validate AI models by providing insights into their inner workings [41].

Another important XAI method in computer vision is the use of interpretable models like decision trees or linear models, which, although less complex, offer greater transparency. Model-agnostic methods, such as Local Interpretable Model-agnostic Explanations (LIME) [37] and SHapley Additive exPlanations (SHAP) [42], are also widely used. LIME creates locally interpretable models to approximate the predictions of more complex models in the vicinity of a specific instance, while SHAP values provide a unified measure of feature importance by considering the contribution of each feature across different model predictions. These XAI methods are crucial for diagnosing model errors, improving model robustness, and ensuring ethical AI deployment by making AI systems more understandable and accountable to human users.

LIME and SHAP, while powerful and popular for explaining the predictions of ML models, face significant challenges when applied to object detection models. These methods typically require access to the class probabilities or objectness scores for the predictions, which are not always available in black-box testing scenarios. Additionally, the random sampling approach used by these methods can lead to inconsistencies in the saliency maps generated. For instance, LIME and SHAP might assign similar importance to both relevant and irrelevant pixels because they do not account for the spatial structure of images and the dependencies between neighbouring pixels. This random masking can result in noisy explanations where irrelevant pixels are marked as important, thus distorting the interpretation of the model's predictions and reducing the overall reliability of the explanations.

Various explanation methods have been developed for computer vision applications [43-47]. However, most of these methods focus on white-box explanations, necessitating access to the ML model's internals. Few studies have addressed black-box explanations for DL models in image processing tasks. One such method is Randomized Input Sampling for Explanation (RISE) [39], which estimates a saliency map by probing the object classification model using randomized masking of the input image. RISE calculates importance scores for pixels by measuring the difference in class probabilities before and after masking the image. Detector Randomized Input Sampling for Explanation (D-RISE) [48] extends RISE's image masking strategy to provide an attribution method for explaining object detection models.

However, RISE and D-RISE face significant challenges. They require probability scores over classes and an objectness score for every bounding box to estimate the saliency map, limiting their utility when only bounding box coordinates are available. Additionally, the random masking process can result in equal importance being assigned to both relevant and irrelevant pixels. This occurs because some pixels are masked more frequently than others, leading to a saliency map where some elements have higher scores due to appearing in more masks. Furthermore, random masks may include both relevant and irrelevant pixels to a detected object, causing irrelevant pixels to appear as noise in the final saliency map. This is because only masking relevant pixels affects the model's output, whereas masking irrelevant pixels does not. Consequently, irrelevant pixels can distort the final interpretation.

**1.3. Contribution of this paper**

In this paper, we address the aforementioned challenges by proposing a new explanation method called Black-box Object Detection Explanation by Masking (BODEM). BODEM is a model-agnostic explanation method that can be applied to any object detection system, regardless of the underlying ML model. Importantly, BODEM does not require access to class probabilities or objectness scores, making it suitable for black-box explanation scenarios where only bounding box predictions are available.

BODEM consists of three main stages: hierarchical random mask generation, model inquiry, and saliency estimation. In the mask generation phase, coarse-grained masks are created at higher levels to identify the most salient regions of an object within the input image. Fine-grained masks are then used at lower levels to refine the saliency estimation within these important regions. During the model inquiry step, the detection model is probed with the masked images to observe changes in its output when certain input information is missing. Finally, based on the differences in the model's output after masking the input, the explanation method computes a saliency map that estimates the importance of pixels for each detected object. This process of masking, inquiry, and saliency estimation is repeated in several iterations, starting from higher levels of the masking hierarchy and progressing to lower levels. The final saliency map is used to visualize a heatmap, illustrating the importance of areas within the image to specific predictions.

The Generation of Saliency Maps based on Hierarchical Masking (GSM-HM) [49] is a related work that generates saliency maps for object detection models using a hierarchical masking strategy. However, GSM-HM relies on objectness scores for saliency estimation, which may not be available in black-box testing and explainability scenarios. In contrast, BODEM only requires object coordinates, making it more suitable for generating black-box explanations. Additionally, the method of incorporating saliency values from lower levels of the masking hierarchy differs between GSM-HM and BODEM.

We conducted extensive experiments on three object detection tasks—user interface control detection, airplane detection, and vehicle detection—using three popular object detection DL models: YOLO [17], R-CNN [13], and SSD [18]. We employed three quantitative measures to assess the accuracy and stability of the explanations. The experimental results demonstrated that BODEM outperforms D-RISE and LIME across all the three metrics. Our findings indicate that BODEM can effectively and reliably produce explanations that reveal the importance of different parts of an image for specific detections. By analyzing the explanations generated by BODEM, we show that the decisions of the detection models can be understood, helping users to grasp the behavior of object detectors. Furthermore, BODEM's explanations reveal certain vulnerabilities of the object detectors for specific types of objects.

## 2. Model-agnostic explanation method

In this section, we give a detailed description of the BODEM explanation method. As already explained, BODEM consists of three main stages, i.e. 1) mask generation, 2) model inquiry, and 3) saliency estimation. **Figure 1** illustrates the overall architecture of the explanation method. In the next subsections, we describe these three phases in detail.

**Problem formulation:** Let $I$ be an input image with dimensions $W \times H$. Given an object detector $f(I) \rightarrow O$, where $O=\{o_1, o_2, \ldots, o_N\}$ is the set of objects detected by $f$, and each object $o_n=(x_1, y_1, x_2, y_2)$ is represented by its bounding box coordinates in a 2D space, our objective is to generate a saliency map $SM_n$ for each detected object. The saliency map has the same dimensions as the input image ($W \times H$) and assigns values that indicate the importance of each pixel in relation to the target object. Our explanation model addresses this problem in a black-box manner, meaning it does not require access to the architecture, loss function, gradients, weights, or output probabilities of the detection model. Additionally, it does not rely on class probabilities or objectness scores for each bounding box detected by the object detector.

A central aspect of our explanation method is the use of a random mask generation technique combined with a hierarchical masking algorithm. This process begins with coarse-grained masks to identify the most salient regions of the image and progressively refines these regions with fine-grained masks. This approach generates a final saliency map with smoother salient areas. At each level of the masking hierarchy, only regions that had non-zero saliency values in the previous level are refined further. By incorporating controlled randomness into the mask generation process, we reduce the noise in the final saliency map.

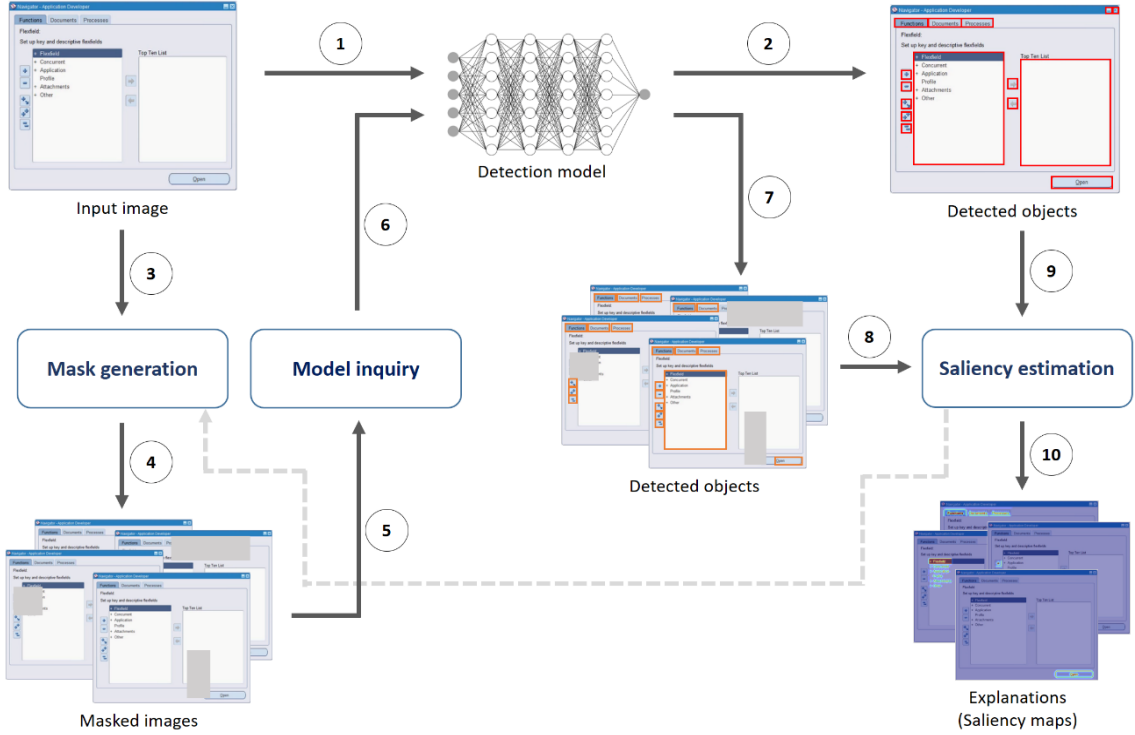

**Figure 1.** The overall architecture of BODEM explanation method. Final explanations are generated through several steps: 1) the input image is fed into the detection ML model, 2) the detection model predicts bounding boxes that specify the detected objects, 3) the mask generation module receives the input image, 4) it then generates the masked images through a hierarchical masking process, 5) the model inquiry module receives the masked images, 6) it then passes the masked images to the detection model, 7) new predictions are generated for the masked images by the detection model, 8) new bounding boxes are given to the saliency estimation module, 9) the original bounding boxes detected by the ML models are also received by the saliency estimation module, and 10) the final explanations are generated as heatmaps through estimating the saliency of pixels. There is also a connection between the saliency estimation and mask generation modules to control the iterative hierarchical random mask generation process based on the saliency values.

### 2.1. Mask generation

As discussed in Section 1, a common approach for providing black-box explanations of image processing models involves masking different parts of the input image and observing how these changes affect the model's output. Techniques such as RISE [39] and D-RISE [48] employ random masking strategies. However, these methods have notable limitations. The primary issue is that random masks may simultaneously cover both relevant and irrelevant pixels. While only relevant pixels influence the model's output, irrelevant pixels can receive importance scores similar to those of the relevant ones, resulting in noise within the saliency map.

To overcome this challenge, we propose a novel masking technique that integrates random and hierarchical masking strategies. In the initial levels, the method identifies the most salient regions of the image using coarse-grained masks. This information is then used to guide the mask generation process in subsequent levels, where fine-grained masks are applied to refine the saliency map continuously. This hierarchical approach ensures that the final saliency map highlights the most important regions with minimal noise.

Given an input image $I$ and a detected object $o_n=(x_1, y_1, x_2, y_2)$, the explanation method begins at level $l=1$ by dividing $I$ into a set of blocks $B=\{b_1, b_2, \ldots, b_P\}$, where each block has dimensions $K \times K$ pixels. Masks $M=\{m_1, m_2, \ldots, m_Q\}$ are then generated by selecting blocks from $B$ and setting the pixel values within those blocks to zero. Specifically, a block $b_p$ is included in a mask $m_q$ if $m_q(b_p)=1$, meaning its pixel values are set to zero. Conversely, if $m_q(b_p)=0$, the block $b_p$ is not masked and its pixel values remain unchanged. The mask generation at each level is guided by information from the saliency estimates obtained in the previous level.

Initially, a set of Candidate Seed (*CS*) blocks is created, including all blocks that received a non-zero saliency value in the previous level. For the first level, where no prior saliency information is available, all blocks are included in the *CS* set. A mask is then created by randomly selecting a seed block from *CS* and masking it, along with 50 percent of the neighboring blocks within a distance of $l$ blocks from the seed. The selection probability for a candidate block as the seed is weighted based on its saliency value from the previous level. In contrast, the neighboring blocks are selected randomly with equal probability. This approach takes advantage of the fact that salient pixels often form contiguous regions of varying sizes. At level $l=1$, all candidate blocks are equally likely to be chosen as the seed for mask generation.

When a block is selected as the seed, it is removed from *CS*. The process of selecting seed and neighboring blocks to generate masks continues until certain termination conditions are met. At $l=1$, because the blocks are larger and the *CS* set is relatively small, the mask generation process terminates when *CS* becomes empty. For levels greater than one, mask generation continues until either: 1) all blocks with a non-zero saliency value from the previous level appear at least once in a mask in the current level, or 2) the *CS* set becomes empty. The first condition ensures that the number of generated masks does not become excessively large, as it is likely that all relevant blocks will be included in at least one mask before *CS* becomes empty at the current level.

In the next level, the width and height of the blocks are halved, reducing their size. For instance, at $l=2$, each block has dimensions $\frac{K}{2} \times \frac{K}{2}$. Figure 2 illustrates examples of blocks and their neighbors (with a distance of $l$ blocks) at levels one and two, with seed and neighboring blocks shown in black and gray, respectively. In Figure 1(a), at level one, blocks within one block distance from the seed are considered neighbors, and 50 percent of these, or four blocks, are randomly selected to form a mask. In Figure 1(b), at level two, blocks within a two-block distance are considered neighbors, and 50 percent of these, or 12 blocks, are chosen to create a mask. As shown, the block size decreases from level *l* to *l*+1. Consequently, the area from which blocks are selected and masked becomes smaller in the next level, as indicated by the red and blue dotted rectangles in Figure 2. Thus, while the number of masked blocks increases, the size of the blocks decreases, and the masked area becomes more focused as the levels progress.

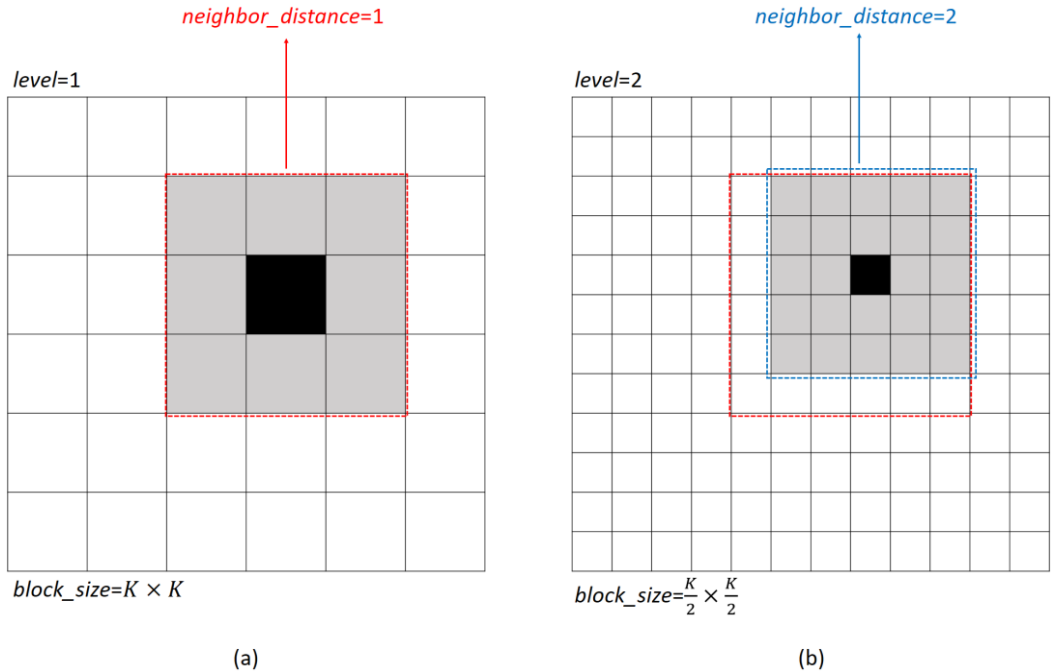

**Figure 2.** Examples of seed blocks and their neighbor blocks in two different levels of the mask generation step. Seed and neighbor blocks are shown by black and gray squares, respectively.

Algorithm 1 provides the pseudocode for mask generation at level *l* in the BODEM explanation method. Initially, the new block size is computed by halving the width and height of the current blocks, and the image pixels are divided into blocks accordingly (lines 3-8). Next, blocks that received non-zero saliency values in the previous level are added to the set of candidate seeds (lines 9-14). The mask generation process then proceeds with selecting seed blocks, selecting neighboring blocks, and adding these to the list of generated masks (lines 15-19). Following this, the selected seed block is removed from the candidate list, and the termination conditions are evaluated. If the conditions are satisfied, the mask generation process concludes at level *l*, and the list of generated masks is returned (lines 20-25).

**Algorithm 1.** The mask generation procedure at level $l$ used by our BODEM explanation method.

1: **Inputs:** input image $I$, detected object $O_n$ within $I$, saliency map $SM_n$, block size in previous level $BS^{l-1}=K \times K$

2: **Output:** set of masks $M^l$ at level $l$

3: **if** $l=1$ **then**

4:     $BS^l = K \times K$ (there is no previous level at level one)

5: **else if** $l>1$ **then**

6:     $BS^l = \frac{BS^{l-1}}{2} = \frac{K}{2} \times \frac{K}{2}$

7: **end if**

8: Divide pixels within $I$ into a set of blocks $B^l=\{b_1, b_2, …, b_P\}$ with block size $BS^l$

9: Create a set of candidate seeds $CS=\emptyset$

10: Create a set of masks $M^l=\emptyset$

11: **for** every block $b_p \in B^l$ **do**:

12:     **if** $l=1$ **then** add $b_p$ to $CS$

13:     **else if** $l>1$ and $SM_n^{l-1}[b_p]$ has any non-zero values **then** add $b_p$ to $CS$

14: **end for**

15: **while** termination conditions are not met **do**:

16:     Randomly select a block $b_p$ from $CS$ as a seed with a probability weighted on saliency $SM_n^{l-1}[b_p]$

17:     Randomly select 50 percent of neighbors of $b_p$ whose distance with $b_p$ is not larger than $l$ blocks

18:     Create a mask $m_q$ where block $b_p$ and the selected neighboring blocks are masked

19:     Add $m_q$ to $M^l$

20:     Remove $b_p$ from $CS$

21:     Check termination conditions:

22:         **if** $l=1$ **then** termination condition is: $CS$ becomes empty

23:         **if** $l>1$ **then** termination conditions are: $CS$ becomes empty or every block $b_p \in B^l$ with $SM_n^{l-1}[b_p]>0$ appears at least in one mask

24: **end while**

25: **return** $M^l$

## 2.2. Model inquiry

The model inquiry module serves as a mediator between the explanation model and the object detection model. It receives a masked image, forwards it to the object detection model, obtains the new bounding boxes detected by the object detector, and then sends these results to the next step, i.e. the saliency estimation module. This inquiry module is model-agnostic, meaning it does not require any knowledge of the neural network architecture, loss function, activation functions, connection weights, or hyperparameters of the underlying machine learning model. It simply sends requests to the detection model and receives the bounding box coordinates predicted by the model. Consequently, this explanation method is capable of explaining any object detection system, regardless of the underlying ML model, making it particularly suitable for black-box software testing scenarios.

## 2.3. Saliency estimation

The saliency estimation module receives the original object coordinates $o_n=(x_1, y_1, x_2, y_2)$ generated by the object detector, along with the new object coordinates $o'_n=(x'_1, y'_1, x'_2, y'_2)$ predicted for the masked image. Its primary function is to estimate the importance of pixels within the input image by assessing the difference between the original and new predictions. It then updates the saliency map $SM_n$, which represents the significance of various parts of the image to the object $o_n$. Initially, the saliency map $SM_n$ is filled with zero values. It is important to note that if multiple objects are detected within the masked image, the one closest to $o_n$ is considered as $o'_n$.

At level $l$, given the original object coordinates $o_n=(x_1, y_1, x_2, y_2)$ detected within the image $I$, a mask $m_q$ applied to the image, the new object coordinates $o'_n=(x'_1, y'_1, x'_2, y'_2)$ detected within the masked image $I'$, and a saliency map $SM_n$, the saliency estimation module first computes the similarity between $o_n$ and $o'_n$ using Intersection Over Union (IOU), as follows:

$$Similarity(o_n, o'_n) = IOU(o_n, o'_n) = \frac{|o_n \cap o'_n|}{|o_n \cup o'_n|} \qquad (1)$$

The computed similarity ranges from 0 to 1. If the saliency estimation module has access to objectness scores or class probabilities, these can also be incorporated into the similarity measure in Equation (1). However, in this paper, we focus on an extreme black-box scenario where only object coordinates are available. Other outputs from object detection models can be easily included in Equation (1) to measure the similarity between two objects.

A similarity value close to one indicates little or no difference between $o_n$ and $o'_n$. Therefore, the blocks that were masked within $I'$ had little or no significance to the target object. Conversely, a similarity value close to zero signifies high importance of the masked blocks to the target object, as the absence of these pixels caused the object detector to predict an incorrect bounding box. Given on the similarity value, an importance score *IS* is computed for each block $b_p$ covered by the mask $m_q$, as follows:

$$IS(b_p) = 1 - Similarity(o_n, o'_n) \qquad (2)$$

After calculating an importance score for each block that was masked in an image at level *l*, an overall importance score *OIS* is computed for every block $b_p$. This is done by averaging all the importance scores across all images where $b_p$ was masked at this level, as follows:

$$OIS(b_p) = \frac{\sum_{m_q \in M^l \mid m_q(b_p)=1} IS(b_p)}{\sum_{m_q \in M^l} m_q(bp) = 1} \qquad (3)$$

where $M^l$ is the set of all masks at level *l*. In Equation (3), the numerator sums all the importance scores for the block $b_p$ across every mask where $b_p$ was masked at level *l*, while the denominator counts the number of masks in which $b_p$ was masked at this level. Essentially, the larger the $OIS(b_p)$, the greater the importance of the block $b_p$ for detecting the target object.

With the overall importance score computed for each block, the saliency map $SM_n$ is updated to reflect the saliency of blocks at level *l*. It is important to note that only blocks with non-zero saliency values at level *l*-1 are considered in the saliency estimation phase at level *l*. If a block was assigned a zero saliency value at level *l*-1, its saliency value will remain zero at level *l*. Additionally, if a block had a non-zero saliency value at level *l*-1 but receives an overall importance score of zero at level *l*, it retains its saliency value from level *l*-1, but this value is reduced to penalize the block for not being salient at level *l*.

The new saliency value of a block is calculated by combining its saliency from level *l*-1 with the overall importance score at level *l*, as follows:

$$SM_n^l[b_p] = \begin{cases} (\alpha)SM_n^{l-1}[b_p] + (1-\alpha)OIS(b_p), & \exists m_q \in M^l \mid m_q(b_p) = 1 \text{ and } OIS(b_p) \neq 0 \\ (\beta)SM_n^{l-1}[b_p], & otherwise \end{cases} \qquad (4)$$

where *α* is a hyperparameter that determines the extent to which the block $b_p$ is influenced by its saliency value at level *l*-1. A higher value of *α* increases the influence of the saliency value from level *l*-1 on the saliency of the block at level *l*. Another hyperparameter, *β*, controls the penalty applied to the block $b_p$ if it does not obtain an overall importance score greater than zero at level *l*. A lower value of *β* results in a greater penalty, causing the block to inherit less of its saliency value from level *l*-1. Both hyperparameters *α* and *β* have values within the range [0, 1].

The process of mask generation, model inquiry, and saliency estimation continues iteratively until the final level is reached. The saliency map produced at the last level contains the most detailed saliency values, refined through several iterations of both coarse-grained and fine-grained masking and saliency estimation. Thus, the saliency map from the last level is used by our BODEM method as the final explanation for the target object.

Algorithm 2 provides the pseudocode for saliency estimation at level $l$ in our BODEM explanation method. Initially, the similarity between the objects in the original and masked images is calculated, and an importance score is estimated for the masked blocks (lines 3-10). Next, an overall importance score is computed for these blocks. The saliency map is then updated for blocks with non-zero saliency values at level $l$-1, and the updated saliency map is returned (lines 11-17).

**Algorithm 2.** The saliency estimation procedure at level $l$ used by our BODEM explanation method.

1: **Inputs:** detected object $O_n$ within image $I$, a set of masks $M^l=\{m_1, m_2, \ldots, m_Q\}$ applied to $I$, set of detected objects $O'_n=\{O'^1_n, O'^2_n, \ldots, O'^Q_n\}$ within masked images $I'_1, I'_2, \ldots, I'_q$ such that $O'^q_n$ was detected within $I'_q$, set of blocks $B^l=\{b_1, b_2, \ldots, b_P\}$ at level $l$, saliency map $SM_n^{l-1}$ from previous level

2: **Outputs:** saliency map $SM_n^l$ after updating in current level

3: **for** every object $O'^q_n$ such that it was detected in image $I'_q$ masked by $m_q$ **do**:

4:    Compute *Similarity*($O_n$, $O'^q_n$) using Equation (1)

5:    **for** every $b_p \in B^l$ **do**:

6:       **if** $m_q(b_p)=1$ **then**:

7:          Compute and store an importance score *IS*($b_p$) using Equation (2)

8:       **end if**

9:    **end for**

10: **end for**

11: **for** every $b_p \in B^l$ **do**:

12:    Compute an overall importance score *OIS*($b_p$) using equation (3) and importance scores computed in line 7

13:    **if** $SM_n^{l-1}[b_p]>0$ **then**:

14:       Update $SM_n^l[b_p]$ using Equation (4)

15:    **end if**

16: **end for**

17: **return** $SM_n^l$

Figure 3 shows an input image with a detected button user interface control, along with the saliency maps generated by our BODEM explanation method at various levels. At the coarse-grained levels, such as $l=1$ and $l=2$, where blocks are larger, the explanation method identifies the salient regions of the image. However, at these levels, the saliency map lacks accuracy, and the important regions do not have smooth boundaries. On the other hand, at the fine-grained levels, such as $l=6$, where blocks are smaller, the salient regions are more precisely refined. Consequently, the final saliency map represents the salient regions, objects, lines, and other features more accurately, with smoother boundaries.

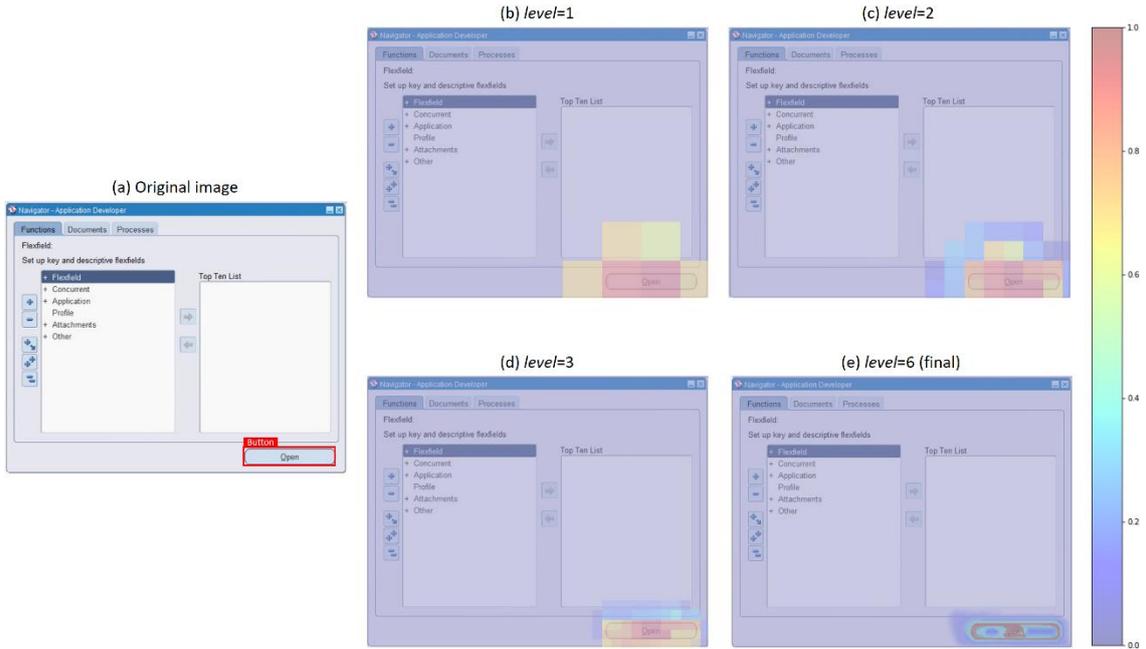

**Figure 3.** An input image and saliency maps generated by our BODEM explanation method at different levels for a detected button user interface control. In this example, the block size is 128×128, 64×64, 32×32, 16×16, 8×8, and 4×4 pixels at level 1, 2, 3, 4, 5, and 6, respectively.

## 3. Experimental results

In this section, we first describe the datasets, evaluation metrics, and object detection models that we utilized in our experiments. We then present the object detection test results, as well as explanations produced by our BODEM explanator. Moreover, we give examples where explanations can help to analyze which parts of objects are more important to the object detectors. In the experiments presented in this section, we used six levels of masks with the following sizes in the BODEM explanation method: 128×128, 64×64, 32×32, 16×16, 8×8, and 4×4. The experimental details, data, and source codes can be accessed at https://github.com/mmoradi-iut/BODEM.

We compare our BODEM explanation method against two baselines, i.e. D-RISE and LIME. As we explained in Section 1.2, D-RISE generates random masks, applies the masks to the input image, probes the detection model with the masked images, and computes a saliency map by measuring the difference in class probabilities and objectness scores before and after masking. LIME depends on random masking, however, it adopts a different importance estimation strategy. LIME creates several perturbed versions of the original image by dividing the image into superpixels (contiguous regions of similar pixels) and randomly turning off some superpixels in each perturbed version. It then uses the object detection model to get predictions for each perturbed image, and constructs a new data set where each sample is a perturbed image, and the corresponding features are the binary indicators of whether each superpixel is turned on or off. The target variable is the prediction made by the detection model for that perturbed image. LIME trains a linear regression model on this new dataset. We used the coefficients of this model as importance scores for the superpixels and visualized them on a heatmap as the final explanation.

We used the source codes provided by the respective authors to implement the D-RISE and LIME baselines for the performance evaluation experiments. We also used the same settings described in the respective papers for the baselines.

### 3.1. Datasets

**User interface control detection**: Our main focus for developing the explanation method was on the user interface control detection task for automated software testing. This is a private dataset, which contains 16,155 images annotated for detection and classification of 18 types of user interface controls. The images are digital screenshots taken from desktop, Software as a Service (SaaS), and mobile applications. We split the dataset into a train and development set containing 14,155 images, and a test set containing 2,000 images.

**Airplane detection**[1]: This public dataset contains 733 aerial images annotated for detecting airplanes. We split the dataset into a train and development set with 650 images, and a test set with 83 images. We chose to use this small dataset in order to investigate how the BODEM explanation method can be effective in scenarios where there are not many images to train an object detection model.

**Vehicle detection**: Common Objects in Context (COCO) dataset [50] is a large dataset of more than 160,000 images annotated for object detection and classification of 80 object categories. We created a subset of 8,000 images from COCO for our vehicle detection task by randomly choosing 2,000 images from each one of the categories "CAR", "BUS", "TRUCK", and "MOTORCYCLE". The train and development set of this new dataset has 7,000 images. The test set contains 1,000 images.

---

[1] https://www.kaggle.com/datasets/airbusgeo/airbus-aircrafts-sample-dataset

### 3.2. Evaluation metrics

We used three different evaluation metrics to objectively evaluate the effectiveness of our explanation method against D-RISE. The evaluation metrics are as follows:

**Deletion:** Pixels of the original image are deleted one by one in descending order of their saliency value, and the difference between the detection results of the original image and the image with deleted pixels is computed as the IOU between the objects detected within the images. We use the mean detection difference Area Under the Curve (AUC) across all the images in the test set as the final deletion measure to evaluate the explanation methods.

**Insertion:** First, a version of the original image is created where pixels of the saliency area are deleted. Then, pixels are filled one by one into the image in descending order of their saliency value, and the difference between the detection results of the original image and the image filled with salient pixels is computed as the IOU between the objects detected within the images. We use the mean insertion difference AUC across all the images in the test set as the final insertion measure to evaluate the explanation methods.

**Convergence:** This metric estimates the ability of methods in generating stable saliency maps for the same object. We measure the convergence by calculating the difference between three saliency maps resulted from running the explanation method three times (with the same experimental settings) on the same input image. The Euclidean distance is used to calculate the difference between saliency maps, as follows:

$$Convergence = \frac{||SM_1 - SM_2|| + ||SM_1 - SM_3|| + ||SM_2 - SM_3||}{3} \quad (5)$$

where $||SM_1 - SM_2||$ denotes the Euclidean distance between two saliency maps $SM_1$ and $SM_2$. A smaller Euclidean distance refers to a better convergence, which subsequently means more stable saliency maps generated by the explanation method.

### 3.3. Object detection models

Object detection models can be generally divided into one-stage and two-stage methods. Two-stage detectors break down the problem into two steps, i.e. 1) detecting region proposals, which possibly contain an instance of an object of interest, and 2) classifying those regions with respect to the probability of an object appearing within a region. On the other hand, one-stage detectors utilize end-to-end neural networks to predict bounding boxes and class probabilities of detected objects all at once. We used both one-stage and two-stage object detection models in our experiments.

**YOLO** [17]: It is a one-stage object detection algorithm that divides the input image into $N$ grids with equal size, and then detects and localizes objects within each grid. In order to handle overlapping bounding boxes detected within different grids, YOLO utilizes Non-Maximal Suppression (NMS), which is a technique to filter grids and select regions with the highest probability of containing an object of interest. Due to its one-stage detection strategy, YOLO can perform much faster than popular two-stage detectors such as R-CNN and Fast R-CNN. That is why it is widely used for real-time object detection. Inspired by the GoogleNet architecture, YOLO composes of 24 convolutional blocks followed by two fully-connected layers. We used YOLO-v5 in our experiments, and trained it on the user interface control detection dataset.

**R-CNN** [13]: This object detector uses selective search to extract region proposals, which are then fed to a CNN that classifies the regions as they contain a target object or not. As the backbone of this object detector, we used a VGG-16 model, which has a CNN with 16 convolutional layers and 134 million parameters. This CNN was already pretrained on more than one million images from the ImageNet dataset[2]. One flatten layer, two fully-connected dense layers, and a softmax layer for the final classification were added to the backbone model. We freezed all layers of the backbone except the last three convolutional layers. We then fine-tuned the remaining layers and our custom layers on the airplane detection dataset.

**SSD** [18]: It is another one-stage detection model that is composed of two components, i.e. a backbone model, and SSD head. A pretrained model is usually utilized as the backbone to serve as a feature extractor. SSD head is usually formed by several convolutional layers on top of the backbone. These additional layers are specifically trained for the object detection task at hand. In our experiments, we used a variant of SSD that has ResNet-101, which were pretrained on ImageNet, with around 44 million parameters as the backbone. Six more convolutional layers were added on top of the backbone to form the SSD head and train it for our object detection task. We fine-tuned the SSD model on the vehicle detection dataset.

---

[2] https://www.image-net.org/

### 3.4. Hyperparameter tuning

As we explained in Section 2.3, there are two hyperparameters that control how the saliency map is updated at level *l* using the saliency values estimated at level *l*-1. The hyperparameter $\alpha$ controls how much saliency a block inherits from its saliency value from level *l*-1. Smaller values of this hyperparameter give a higher weight to the saliency score estimated at level *l*, while larger values assign a higher weight to the saliency value estimated at level *l*-1. The hyperparameter $\beta$ controls how much a block is penalized if it gets a zero saliency value at level *l*. Smaller values of this hyperparameter give larger penalties by letting the block inherit only a small proportion of the saliency value it had at level *l*-1. In this way, if $\beta$ has a small value, the saliency value of a block shrinks by a large proportion at the next levels if it gets a zero saliency score at level *l*.

We conducted a set of hyperparameter tuning experiments on the training sets to find optimal values for $\alpha$ and $\beta$, with respect to the accuracy of the explanations and the visual quality of the saliency maps. The results were to a high extent similar across all the three datasets. We observed the highest performance scores and visual quality of saliency maps when $\alpha=0.3$ and $\beta=0.2$. Therefore we used these hyperparameter values in the subsequent experiments on the test sets.

### 3.5. User interface control detection

We trained the YOLO detection model on the user interface control detection dataset. **Table 1** presents the performance scores obtained by the YOLO object detection model on the respective test set. Although these scores are not relevant to the experiments we performed to evaluate the effectiveness of the explanation methods, we present them in the paper only to give readers an idea how well the object detection model performed on this task.

Table 2 reports the performance evaluation scores obtained by our BODEM explanation method, D-RISE, and LIME on the objects detected by the YOLO model on the user interface control detection dataset. As can be seen, BODEM obtained a lower mean deletion AUC than the two baselines. This demonstrates that BODEM performs more accurate than D-RISE and LIME in detecting the salient regions within the images, as deleting salient pixels detected by BODEM results in a more rapid decrease in the detection accuracy in comparison to the other explanators. Moreover, a higher mean insertion AUC is reported for BODEM. It again refers to the superior ability of BODEM in comparison to D-RISE and LIME in detecting the most salient image areas, as inserting the salient pixels identified by BODEM into the images led to a quicker increase in the detection accuracy. Our BODEM explanator also obtained a better mean convergence, demonstrating its ability to generate more stable saliency maps. The hierarchical masking strategy has the most contribution to the stability of saliency maps by controlling the randomness and limiting random masks to the most salient regions instead of the whole image.

**Table 1.** Performance scores obtained by the YOLO object detection model on the user interface control detection test set.

| Class | Precision | Recall | mAP@.5 | mAP@.95 |
|---|---|---|---|---|
| ALL | 0.799 | 0.752 | 0.75 | 0.588 |
| ICON | 0.931 | 0.88 | 0.883 | 0.611 |
| DROPDOWN | 0.886 | 0.907 | 0.904 | 0.748 |
| BUTTON | 0.893 | 0.861 | 0.886 | 0.773 |
| MENU | 0.836 | 0.512 | 0.523 | 0.4 |
| INPUT | 0.913 | 0.748 | 0.758 | 0.663 |
| LIST | 0.674 | 0.709 | 0.636 | 0.481 |
| TABBAR | 0.905 | 0.582 | 0.645 | 0.562 |
| TABLE | 0.815 | 0.862 | 0.825 | 0.747 |
| RADIO_SELECTED | 0.916 | 0.916 | 0.946 | 0.68 |
| RADIO_UNSELECTED | 0.856 | 0.957 | 0.918 | 0.688 |
| CHECKBOX_UNCHECKED | 0.891 | 0.94 | 0.925 | 0.654 |
| CHECKBOX_CHECKED | 0.905 | 0.887 | 0.904 | 0.591 |
| TREE | 0.77 | 0.769 | 0.749 | 0.63 |

**Table 2.** The results of performance evaluation experiments obtained by the BODEM, D-RISE, and LIME explanation methods on the objects detected by the YOLO model on the user interface control detection dataset. The best result in each column is shown in underlined face.

| Explanation method | Mean deletion AUC | Mean insertion AUC | Mean convergence |
|---|---|---|---|
| D-RISE | 0.113 | 0.612 | 18.406 |
| LIME | 0.089 | 0.737 | 8.443 |
| BODEM | <u>0.058</u> | <u>0.875</u> | <u>6.051</u> |

Figure 4 shows examples of user interface controls detected by the YOLO object detector, and saliency maps generated by our BODEM explanatory, D-RISE, and LIME. As can be seen, the saliency maps generated by D-RISE somehow identified the salient regions to the detected objects, however, it also identified some irrelevant regions as salient. LIME also identified important superpixels, however, these superpixels may not properly represent important image regions since they were created merely based on continuous pixel similarity and contain irrelevant pixels. Moreover, LIME's importance scores were estimated based on uncontrolled random masking. As the heatmaps show, LIME cannot generate smooth saliency levels, as all pixels in a superpixel receive the same importance score and are not refined further. On the other hand, our BODEM explanator managed to identify the most salient regions more accurately than D-RISE and LIME, with significantly less noise within the saliency maps and smooth saliency levels.

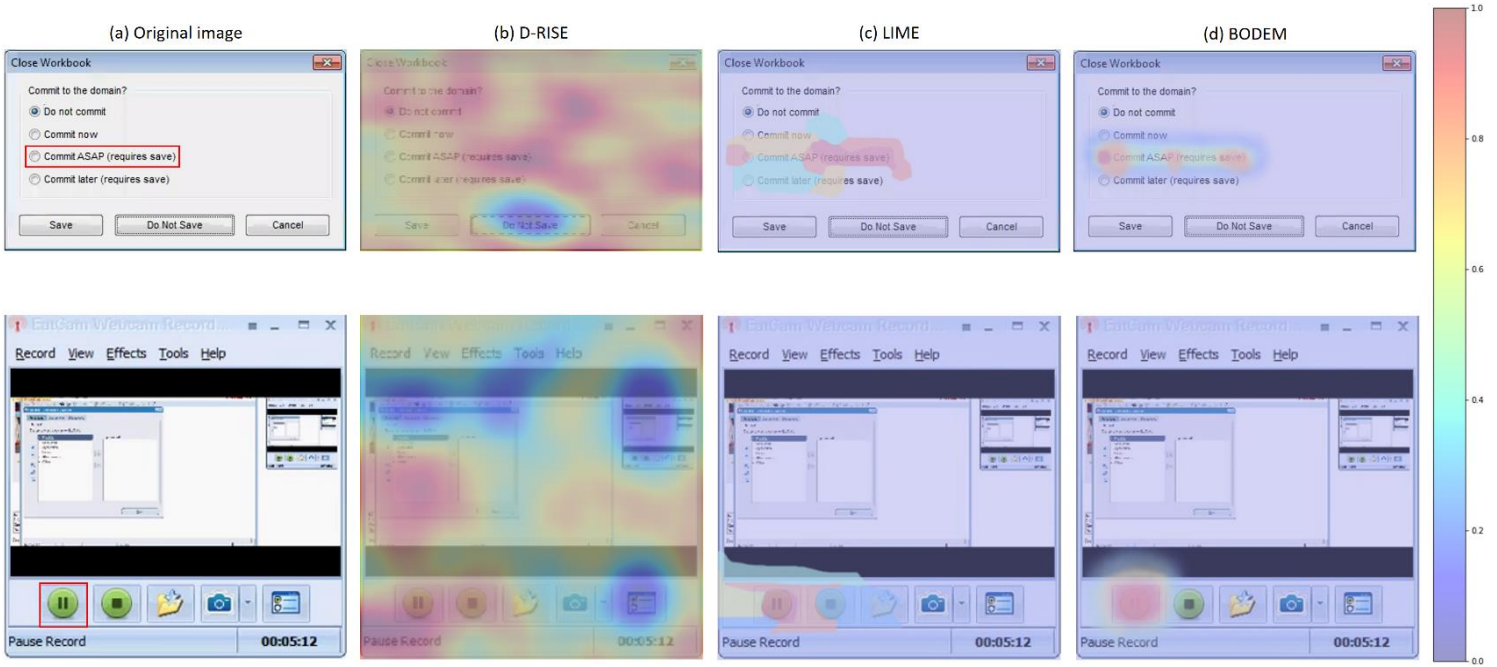

**Figure 4.** User interface controls detected by the YOLO model within two images, and saliency maps generated by our BODEM explanation method, D-RISE, and LIME for the detected objects.

Observing many explanations produced by the BODEM explanator helped us find some patterns that can be useful for inspection and validation of the detection model: 1) lines that specify the borders of controls are very important to the model, 2) icons within BUTTON controls have high impacts on detecting correct boundaries, 3) correct detection of small controls, such as CHECKBOX, is highly dependent on the neighboring areas within the image, such that perturbing small parts of neighboring area can mislead the object detector, 4) texts within controls such as MENU have large effects on the correct predictions, 5) detecting those BUTTON controls that are close to other controls is also influenced by the borders of neighboring controls, in addition to the control's borders, and other similar patterns. These observations convey that explanations generated by BODEM can effectively help us understand the behavior of the detection model.

### 3.6. Airplane detection

We trained the R-CNN detection model on the airplane detection dataset. The model achieved an accuracy of 87.05% on the test set. We utilized our BODEM explanation method to understand which parts of the objects detected by the detection model are more important and have higher impact on the model's decisions.

Table 3 presents the performance evaluation scores obtained by our BODEM explanation method, D-RISE, and LIME on the objects detected by the R-CNN model on the airplane detection dataset. As the results show, our BODEM explanator outperforms D-RISE and LIME with respect to the mean deletion AUC, mean insertion AUC, and mean convergence. This demonstrates the effectiveness of the saliency maps generated by BODEM in identifying the most important regions to the detected objects in the images.

**Table 3.** The results of performance evaluation experiments obtained by the BODEM, D-RISE, and LIME explanation methods on the objects detected by the R-CNN model on the airplane detection dataset. The best result in each column is shown in underlined face.

| Explanation method | Mean deletion AUC | Mean insertion AUC | Mean convergence |
|---|---|---|---|
| D-RISE | 0.128 | 0.597 | 19.512 |
| LIME | 0.093 | 0.712 | 10.085 |
| BODEM | 0.064 | 0.856 | 7.133 |

Figure 5 shows examples of airplanes detected by the R-CNN object detector, and saliency maps generated by our BODEM explanator, D-RISE, and LIME. As can be seen, similar to the user interface control detection dataset, the saliency maps generated by BODEM contain less noise and identify the most important regions more accurately than those generated by D-RISE and LIME. As the saliency maps show, the head, wings and tail of the airplanes have more impact than other parts on detecting the airplanes. This is a common pattern in many objects detected by the model, suggesting that the object detector decides about the bounding boxes by paying more attention to the head, wings, and head of an airplane.

### 3.7. Vehicle detection

We trained the SSD detection model on the vehicle detection dataset. **Table 4** presents the performance scores obtained by the SSD object detection model on the respective test set. Although these scores are not relevant to the experiments we performed to evaluate the effectiveness of the explanation methods, we present them in the paper only to give readers an idea how well the object detection model performed on this task.

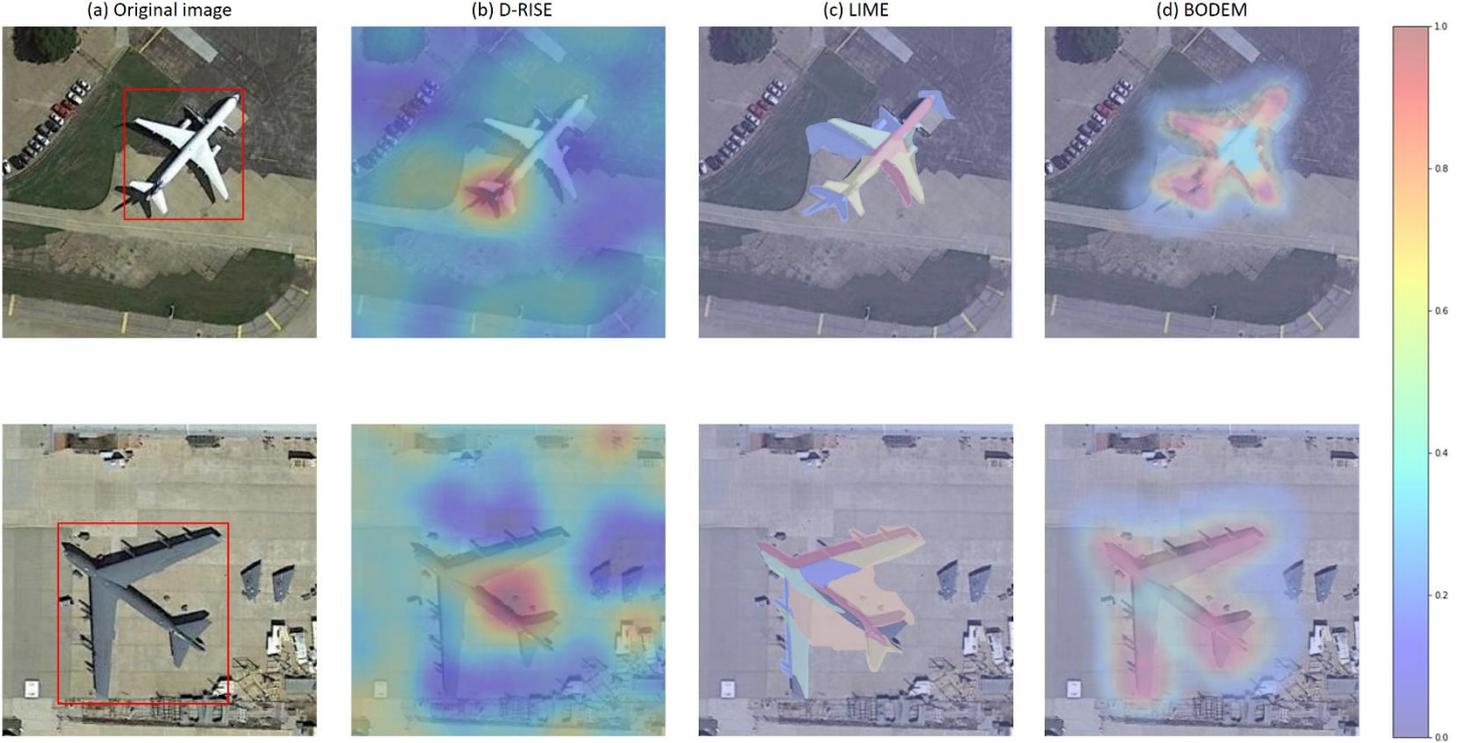

**Figure 5.** Airplanes detected by the R-CNN model within two images, and saliency maps generated by our BODEM explanation method, D-RISE, and LIME for the detected objects.

**Table 4.** The performance scores obtained by the SSD detection model on the vehicle detection test set.

| Class | Precision | Recall | mAP@.7 |
|---|---|---|---|
| ALL | 0.813 | 0.780 | 0.765 |
| CAR | 0.844 | 0.801 | 0.783 |
| BUS | 0.819 | 0.781 | 0.765 |
| TRUCK | 0.785 | 0.752 | 0.742 |
| MOTORCYCLE | 0.803 | 0.788 | 0.771 |

Table 5 reports the performance evaluation scores obtained by our BODEM explanation method, D-RISE, and LIME on the objects detected by the SSD model on the vehicle detection dataset. Similar to the results reported on the other two datasets, BODEM obtained better results than D-RISE and LIME regarding all the metrics. This again demonstrates the effectiveness of our explanation method for identifying the most salient regions to the objects across various detection tasks.

Figure 6 shows examples of vehicles detected by the SSD object detector, and saliency maps generated by our BODEM explanator, D-RISE, and LIME. Similar to the saliency maps generated for the other two tasks, BODEM managed to generate more accurate explanations with less amounts of noise. As the explanations show, the wheels, hood, and boot had higher impact on detecting the car. The wheels and seat were more important for detecting the motorcycle. Observing several explanations, we found similar patterns in the objects detected by the SSD detection model on the vehicle detection dataset. We also found that the object detector paid more attention to the head and wheels for predicting bounding boxes around buses. Another observation is that the head, wheels, and lower parts of the truck objects had a higher importance to the detection model for this particular class.

**Table 5.** The results of performance evaluation experiments obtained by the BODEM, D-RISE, and LIME explanation methods on the objects detected by the SSD model on the vehicle detection dataset. The best result in each column is shown in underlined face.

| Explanation method | Mean deletion AUC | Mean insertion AUC | Mean convergence |
|---|---|---|---|
| D-RISE | 0.137 | 0.605 | 17.381 |
| LIME | 0.102 | 0.728 | 12.055 |
| BODEM | <u>0.069</u> | <u>0.860</u> | <u>6.420</u> |

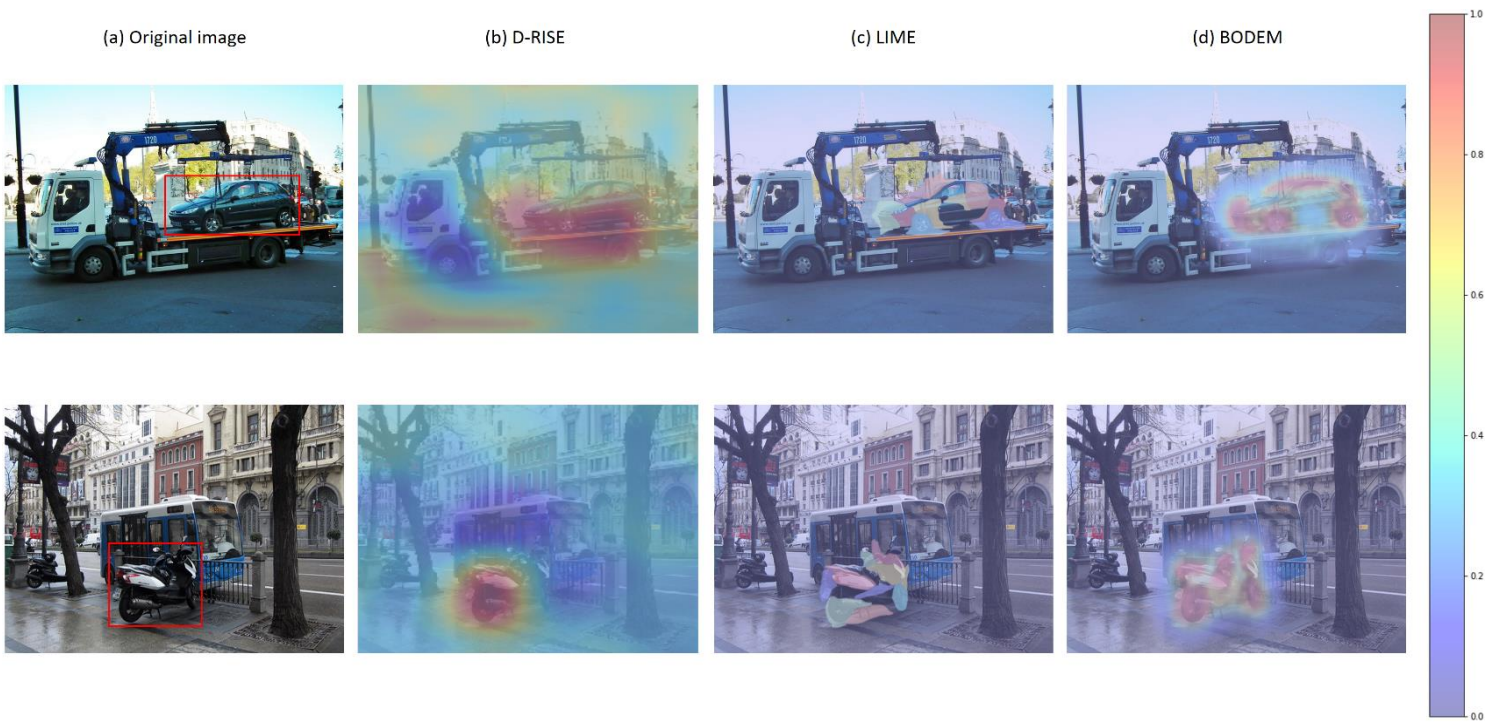

**Figure 6.** Vehicles detected by the SSD model within two images, and saliency maps generated by our BODEM explanation method, D-RISE, and LIME for the detected objects.

## 3.8. Discussion

The superior performance of the BODEM method compared to other existing methods, such as D-RISE and LIME, can be attributed to its novel hierarchical random masking strategy. This approach is particularly effective in generating accurate and stable saliency maps for object detection models, even under black-box testing scenarios.

BODEM addresses the limitations of LIME and D-RISE by introducing a hierarchical random masking strategy specifically designed for object detection models. Unlike LIME and D-RISE, BODEM does not require access to class probabilities or objectness scores, making it more suitable for black-box scenarios where only bounding box predictions are available. BODEM operates by progressively refining the saliency maps through multiple levels of coarse- to fine-grained masks, which helps in accurately identifying the most salient regions associated with the detected objects. This approach minimizes noise and irrelevant pixel inclusion by focusing on regions that are consistently marked as important across different iterations. Consequently, BODEM provides more stable and precise explanations, highlighting the key areas that influence the object detection model's decisions, thereby making it a more effective tool for understanding and validating these models.

The hierarchical random masking strategy is central to BODEM's effectiveness. It starts by generating coarse-grained masks at higher levels, which helps in identifying the most salient regions of an image related to the detected objects. These coarse-grained masks are then refined progressively at lower levels using fine-grained masks. This multi-level approach ensures that the masking process becomes more focused and accurate over iterations, effectively reducing the noise and improving the precision of the saliency maps. By progressively narrowing down the regions of interest, BODEM mitigates the issue of random noise that often affects other methods, such as RISE and D-RISE, which apply masks in a more arbitrary manner.

BODEM incorporates a controlled randomness mechanism in its mask generation process. Instead of generating masks purely randomly, it bases the refinement of masks on the saliency values obtained from previous iterations. This ensures that only the regions consistently identified as important are further examined, which significantly reduces the inclusion of irrelevant or noisy pixels in the final saliency map. This iterative process not only enhances the stability of the generated explanations but also ensures that the final output accurately highlights the most crucial areas contributing to the model's decision.

One of the key theoretical advantages of BODEM is its applicability in black-box scenarios. Unlike methods that require access to internal model parameters, probabilities, or objectness scores, BODEM only relies on the coordinates of the detected objects. This makes it versatile and suitable for a wide range of applications where the internal details of the object detection model are not accessible. The ability to function effectively without internal model data broadens the applicability of BODEM and allows it to be used in more restrictive environments where data privacy or proprietary concerns limit access to model internals.

Furthermore, the above-mentioned aspects of the method also contribute in providing a robust framework that can mitigate different types of noise, e.g. Gaussian, shot, and impulse noise, effectively:

- **Hierarchical random masking strategy:** The model incorporates a hierarchical random masking strategy, which starts with coarse-grained masks to identify the most salient regions and refines them with fine-grained masks in subsequent levels. This hierarchical approach helps in mitigating the impact of noise by focusing on progressively smaller and more relevant regions, reducing the likelihood that noise significantly affects the final saliency map.
- **Controlled randomness in mask generation:** The method employs controlled randomness in mask generation to ensure that only regions with non-zero saliency values from previous levels are refined further. This approach limits the propagation of noisy pixels through the mask hierarchy, effectively filtering out irrelevant noise.
- **Saliency estimation and mask refinement:** By continuously refining the saliency map through multiple levels and iterations, the model can better isolate and highlight truly relevant features while minimizing the influence of noise. This iterative process ensures that only the most significant regions contribute to the final saliency map, thus reducing the impact of Gaussian, shot, and impulse noise.
- **Mask generation based on saliency values:** The mask generation process uses saliency values from previous iterations to control the selection of candidate blocks for masking. This method inherently reduces noise, as it focuses on regions that have been consistently identified as important, rather than randomly selecting pixels which may include noise.

The BODEM method can have significant practical implications, particularly in the domains of software testing, aviation, and transportation, which were the focus of our experimental evaluation. In automated software testing, understanding the decision-making process of object detection models for user interface controls is crucial. BODEM can provide clear explanations for why certain user interface elements were detected, helping developers to debug and improve their systems efficiently. Similarly, in the aviation industry, accurate detection and interpretation of objects such as airplanes are vital for various applications, including aerial surveillance and airport management. By highlighting the most relevant regions that led to the detection of airplanes, BODEM aids in verifying the reliability of these models, ensuring safety and operational efficiency. In transportation, particularly for vehicle detection tasks, BODEM's ability to explain the importance of different image regions can enhance the development and validation of intelligent transportation systems. This can lead to improved traffic management and autonomous vehicle navigation by providing insights into how these systems detect and classify vehicles under different conditions.

### 3.9. Limitations

While the BODEM method demonstrates effectiveness in providing explanations for object detection models, it is essential to acknowledge its limitations. One notable limitation is the computational complexity associated with the hierarchical random masking strategy. Generating multiple masks and performing repeated model inquiries can be computationally intensive, especially for large images or datasets. This complexity could limit the scalability of BODEM in real-time applications or scenarios involving very high-resolution images. Therefore, as a future research direction, optimizing the computational efficiency of the method is crucial. This could involve exploring parallel processing techniques, leveraging hardware accelerations such as GPUs, or developing more efficient algorithms for mask generation and model inquiry.

While our explanation method outperformed the benchmark methods on the airplane detection task regarding the mean insertion, deletion, and convergence metrics with a test set of 83 images, we recognize that the small size of this test set may limit the generalizability of our findings. However, our method's advantage is not limited to a single dataset. We conducted extensive experiments across two other large datasets: user interface control detection and vehicle detection. This diverse evaluation demonstrates the generalizability and robustness of our approach across various object detection tasks. To further validate the generalization capacity of explanation methods, future work will involve testing on larger and more diverse datasets. Expanding the evaluation to include more comprehensive and varied data will help ensure the robustness and scalability of our method. We also plan to explore additional object detection tasks and integrate new datasets to provide a broader assessment of explanation methods' performance.

The general interpretability of saliency maps generated by explanation methods have already been evaluated extensively in previous work [51-54]. However, we acknowledge a user study of the interpretability of the saliency maps generated by BODEM would complement the quantitative experimental results. It was not feasible to conduct an extensive user study within the scope of this paper due to resource and time limitations. Nevertheless, we recognize the importance of user studies in further validating the practical usefulness of our method. As such, we plan to incorporate an extensive user study in future work. This will involve gathering feedback from end-users and domain experts to assess how effectively the saliency maps aid in interpreting and improving object detection models in practical scenarios. We are committed to ensuring that BODEM not only performs well in quantitative evaluations but also meets the needs of users in real-world applications.

## 4. Conclusion

In this paper, we proposed BODEM, a method for explaining the output of object detection models in a black-box manner. Our explanation method utilizes a hierarchical random masking strategy to identify the most important regions to an object within the input image and estimate a saliency map. We conducted extensive experiments on various object detection models and datasets, using different objective evaluation metrics. The results showed that BODEM can be effectively used to generate visual explanations that reveal which parts of images and objects are more important when an object detector makes a decision. The explanations helped us find useful patterns about the behavior of detection models on different tasks and objects. The explanation method does not need to have access to the underlying ML model's internals or any other information about its structure or settings. This makes BODEM a proper choice for explaining and validating the behavior of object detection systems and reveal their vulnerabilities in black-box software testing scenarios. It was already investigated how failure points of natural language processing models can be revealed by injecting small amounts of noise into the input and observing changes in the behavior of the models [55].

While BODEM is specifically designed for explaining object detection models, its underlying principles and hierarchical masking strategy can be generalized to other computer vision tasks, such as:

- **Image classification:** The hierarchical masking strategy can be adapted to explain image classification models by identifying and highlighting the most important regions that contribute to a particular class prediction.
- **Semantic segmentation:** In semantic segmentation tasks, BODEM can be extended to generate saliency maps that explain which regions of the image contributed to the segmentation of specific objects or classes.

- **Instance segmentation:** Similar to object detection, instance segmentation can benefit from BODEM's ability to explain the importance of different regions in detecting and segmenting individual objects within an image.
- **Action recognition in videos:** For tasks involving video data, such as action recognition, BODEM can be extended to analyze temporal sequences and identify key frames or regions that are crucial for recognizing specific actions.

To further validate and enhance the generalization capacity of BODEM, future work will focus on: 1) testing BODEM on larger and more diverse datasets to ensure robustness and scalability across different scenarios, 2) exploring additional computer vision tasks, such as image classification and semantic segmentation, to adapt and apply the hierarchical masking strategy, and 3) investigating the integration of BODEM with real-time applications, such as autonomous driving and video surveillance, to provide timely and interpretable explanations. By addressing these areas, we aim to extend the applicability of BODEM, making it a versatile tool for enhancing the transparency and trustworthiness of AI systems across a wide range of computer vision tasks.

Other future lines of work may include: 1) combining deep learning with symbolic reasoning to provide more interpretable models, 2) integrating human feedback into the learning process to refine and improve the explainability of AI systems, 3) developing methods that can provide explanations using multiple modalities (e.g., text, visualizations, audio) to cater to different user needs and accessibility requirements, 4) for tasks such as video analysis and real-time object detection, developing methods to explain the temporal and spatial dynamics influencing model decisions, 5) researching how explainability techniques can contribute to the robustness and generalization of computer vision models, and 6) Establishing benchmarks and standardized evaluation metrics for explainability in computer vision.